\documentclass[10pt,twocolumn,letterpaper]{article}

\usepackage{cvpr}              

\usepackage[linesnumbered, boxed, ruled]{algorithm2e}
 \RestyleAlgo{ruled}
 \SetCommentSty{mycommfont}
 \usepackage{amsmath}
 \usepackage{multirow}
 \usepackage{pifont}
\usepackage{commath}
\usepackage[dvipsnames]{xcolor}
\usepackage{soul}

\definecolor{my_yellow}{HTML}{FFC500}

\definecolor{cvprblue}{rgb}{0.21,0.49,0.74}
\usepackage[pagebackref,breaklinks,colorlinks,allcolors=cvprblue]{hyperref}

\def\paperID{5103} 
\def\confName{CVPR}
\def\confYear{2026}

\title{Omni-3DEdit: Generalized Versatile 3D Editing in One-Pass}

\author{
Liyi Chen\thanks{Equal contribution. $^\dagger$Corresponding author.},\ Pengfei Wang\footnotemark[1],\ Guowen Zhang,\ Zhiyuan Ma,\ 
Lei Zhang$^\dagger$\ \  \\
The Hong Kong Polytechnic University\ \\
liyi0308.chen@connect.polyu.hk,  cslzhang@comp.polyu.edu.hk
}

\begin{document}
\maketitle
\begin{abstract}
Most instruction-driven 3D editing methods rely on 2D models to guide the explicit and iterative optimization of 3D representations. This paradigm, however, suffers from two primary drawbacks. First, it lacks a universal design of different 3D editing tasks because the explicit manipulation of 3D geometry necessitates task-dependent rules, e.g., 3D appearance editing demands inherent source 3D geometry, while 3D removal alters source geometry. 
Second, the iterative optimization process is highly time-consuming, often requiring thousands of invocations of 2D/3D updating. We present Omni-3DEdit, a unified, learning-based model that generalizes various 3D editing tasks implicitly. 
One key challenge to achieve our goal is the scarcity of paired source-edited multi-view assets for training. To address this issue, we construct a data pipeline, synthesizing a relatively rich number of high-quality paired multi-view editing samples. Subsequently, we adapt the pre-trained generative model SEVA as our backbone by concatenating source view latents along with conditional tokens in sequence space. A dual-stream LoRA module is proposed to disentangle different view cues, largely enhancing our model's representational learning capability. 
As a learning-based model, our model is free of the time-consuming online optimization, and it can complete various 3D editing tasks in one forward pass, reducing the inference time from tens of minutes to approximately two minutes. Extensive experiments demonstrate the effectiveness and efficiency of Omni-3DEdit. 
Code: \url{https://github.com/mt-cly/Omni3DEdit} .

\end{abstract}   
\section{Introduction}

\begin{figure}[t!]
    \centering
    \includegraphics[width=1\linewidth]{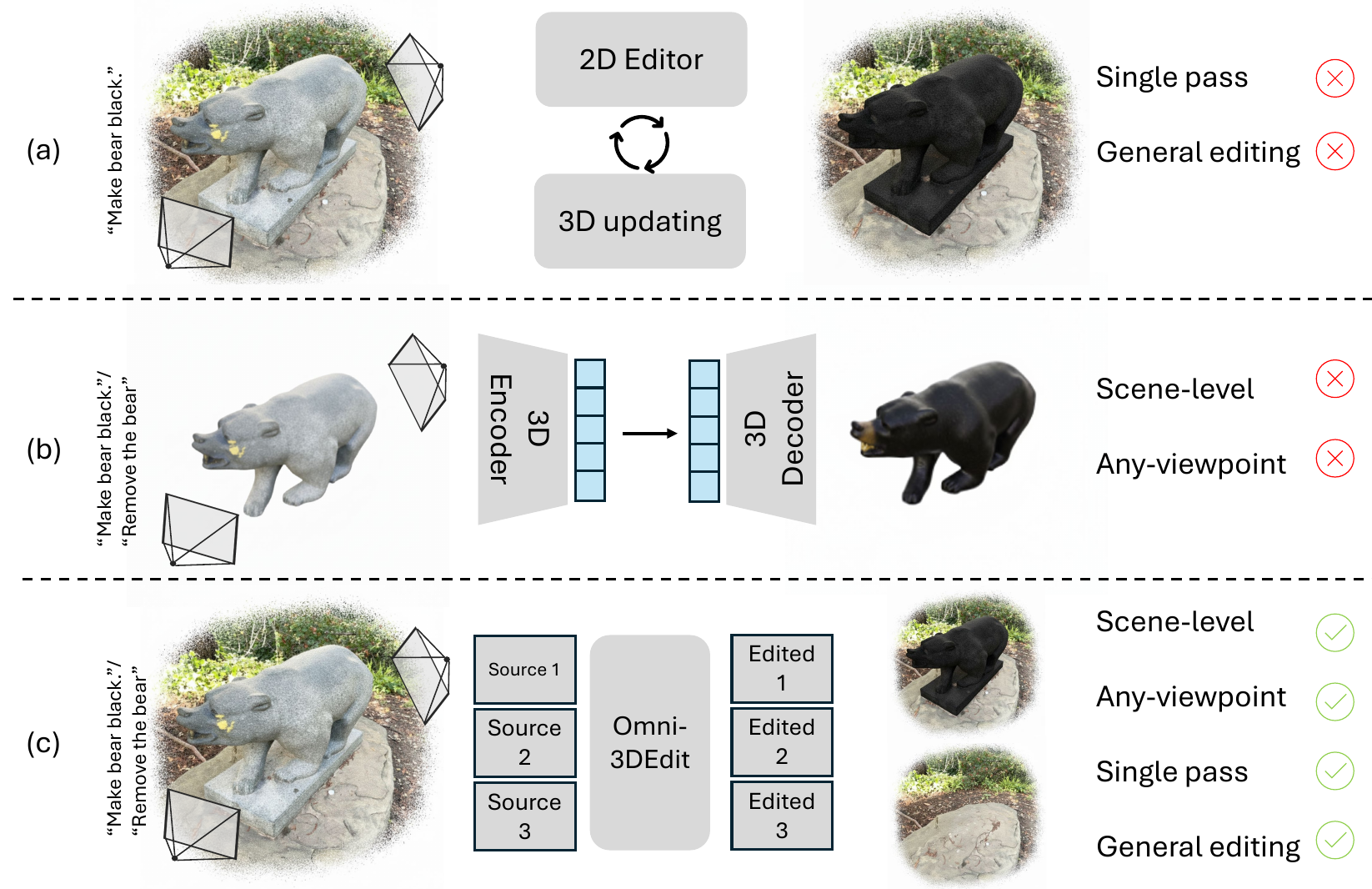}
        \caption{\textbf{Motivation of Omni-3DEdit.} (a) 3D editing via iterative 2D-3D-2D optimization with explicit 3D representation lacks generality and is time-consuming. (b) Performing 3D editing in latent space is hard to handle scene-level assets with arbitrary viewpoints. (c) Our Omni-3DEdit aims to solve these issues in multi-view space to perform fast, general, and consistent editing.  }
    \label{fig:teaser}
    \vspace{-9pt}
\end{figure}

Instruction-based 3D editing aims to edit a given 3D asset according to user's text prompt, encompassing tasks such as altering object appearances~\cite{instructn2n}, adding new objects to specified locations~\cite{mvinpainter}, removing or replacing existing objects~\cite{spinnerf}, \etc. The predominant approach to this challenge involves leveraging 2D vision-language models~\cite{li2024source,wu2026cocoedit,tao2025instantcharacter,zhang2024voxel,yang2023uni,gu2024analogist} to guide the iterative refinement of a 3D representation. However, a fundamental limitation of 2D models is their lack of multi-view consistency. Existing methods~\cite{dge, trame, vcedit, intergsedit,consistdreamer} rely heavily on an iterative 2D-3D-2D optimization loop to alleviate the inconsistencies arising from per-view edits. For instance, to achieve 3D appearance editing, Instruct-N2N~\cite{instructn2n}, DGE~\cite{dge}, and ViCANeRF~\cite{vicanerf} repeatedly sample camera views to compute 2D gradients, updating a NeRF~\cite{nerf} or Gaussian~\cite{3dgs} representation to preserve the original consistent geometry.  In parallel, 3D removal methods~\cite{remove_1,remove_2,spinnerf} often warp foreground masks and employ 2D inpainting models to fill the missing regions, which require explicit 3D updates iteratively as well.

While these methods have achieved notable success in their respective domains, they suffer from two critical drawbacks. First, these approaches are task-specific and lack generality. Appearance editing is heavily reliant on the source 3D geometry, whereas object removal requires masks and often involves large-scale geometric deformations. It is difficult to design a universal iterative rule compatible with diverse editing tasks. Second, multi-round iterative process leads to excessive computation time and can over-smooth texture details. For example, InstructN2N~\cite{instructn2n} requires dozens of minutes for a single appearance edit.

We argue that maintaining and updating an explicit 3D representation, while ensuring consistency, is inherently ill-suited and slow for universal and rapid adaptation to various 3D editing commands. Although recent methods such as Tailor3D~\cite{tailor3d} and CMD~\cite{cmd} have explored editing in a 3D latent space~\cite{lrm,gslrm} to enable a single-pass unified framework, their models are fitted on object-centric datasets (\eg, ObjectVerse~\cite{objaverse}), limiting them to specific camera pose distributions and rendering them incapable of handling general 3D scene inputs.

Instead, in this paper, we introduce \textbf{Omni-3DEdit}, a novel framework that addresses the above-mentioned challenges by performing 3D editing directly in the multi-view latent space, as shown in Fig.~\ref{fig:teaser}. Our model accepts multi-view images of the original 3D asset from arbitrary viewpoints and an editing instruction, and outputs a set of consistently edited multi-view images. Compared to paradigms that operate in explicit 3D space or on object-level 3D latents, our Omni-3DEdit takes advantage of  recent advancements in multi-view generation, 2D editing, and 3D reconstruction.
Specifically, we first employ VGGT~\cite{vggt} to acquire camera cues for the input views, which is crucial for ensuring multi-view consistency. 
We then obtain the reference view by using the recent single-image editor Qwen-Image~\cite{qwenimage} to perform instruction-guided editing on a randomly selected source view. Subsequently, we introduce OmniNet, a model trained to propagate this edited view consistently across the other viewpoints. OmniNet takes the camera poses, the source multi-view images, and the reference image as input to synthesize remaining edited views. {Finally, the resulting  view set can be fed into a reconstruction model (\eg AnySplat~\cite{anysplat}) to obtain edited 3D asset. }

To overcome the scarcity of large-scale paired training data for this task, we adopt a two-pronged strategy. First, we leverage the consistency priors of existing multi-view models to build an offline data synthesis pipeline~\cite{qwenimage,anysplat}, generating paired training data for various tasks, including 3D removal, addition,  and appearance editing. Second, we repurpose the pre-trained multi-view generative model SEVA~\cite{seva} to reduce data dependency. A dual-stream LoRA~\cite{lora} module, composing of \textit{Geometry LoRA} and \textit{Guidance LoRA}, is trained to encode the source and target views, preventing the model from disregarding the crucial geometric priors of the source asset.

In summary, our contributions are threefold. First, we propose Omni-3DEdit, a learning-based framework that operates in the multi-view latent space, enabling single-pass, efficient, and unified editing for diverse, scene-level 3D assets.
Second, we explore and present effective strategies for training a multi-view consistent editing model $\textit{OmniNet}$ in data-constrained scenarios.
Third, extensive experiments demonstrate the superior efficiency and effectiveness of our proposed method in various 3D editing tasks. 
\section{Related Work}

\noindent \textbf{3D Editing in 3D Representation Space.}
Early works aimed to achieve instruction-driven 3D editing by coupling existing 2D multi-modal generation or editing models with 3D representations such as NeRF~\cite{nerf} or Gaussian Splatting~\cite{3dgs}. On one hand, the 2D multi-modal models provide a robust text comprehension interface and editing guidance\cite{li2025syncnoise, chen2025fast}, while the 3D representation ensures that the editing results adhere to 3D geometry. 
To ensure multi-view consistency, most methods iteratively evoke 2D models and 3D representations. 
InstructN2N~\cite{instructn2n}, GaussianEditor~\cite{gaussctrl} and the following works~\cite{clipnerf, nerfart, vcedit,dreameditor} convey view-dependent denoising gradient into 3D representations, iterating thousands of times to achieve appearance editing. 
While object removing and inpainting methods~\cite{spinnerf, mvinpainter,painpainter} share a similar workflow based on 2D inpainters, these methods need to maintain a 3D representation during the editing process and update it in an optimization-based manner. In summary, this category of approaches lacks compatibility across different editing tasks and is highly time-consuming.

\noindent \textbf{3D Editing in Object-level 3D Latent Space.}
Leveraging pre-trained object-level 3D generative models (\eg Shape-E~\cite{shape}, LRM~\cite{lrm}, and GS-LRM~\cite{gslrm}), methods in~\cite{shap-editor, tailor3d, cmd, editp23, nano3d} explore learning an editing mapping within the latent space of the 3D representation. Compared to maintaining an explicit 3D representation, the latent space is easier to be integrated into editing networks, making it possible to learn a direct mapping from the original 3D latent to the edited latent. 
However, these methods require view inputs from specific object-centered camera poses and they are constrained by the limitations of base 3D generation model, which can only handle background-free 3D objects~\cite{objaverse}. As a result, these methods cannot process scene-level editing from arbitrary viewpoints.

\begin{figure*}[t!]
    \centering
    \includegraphics[width=1.0\linewidth]{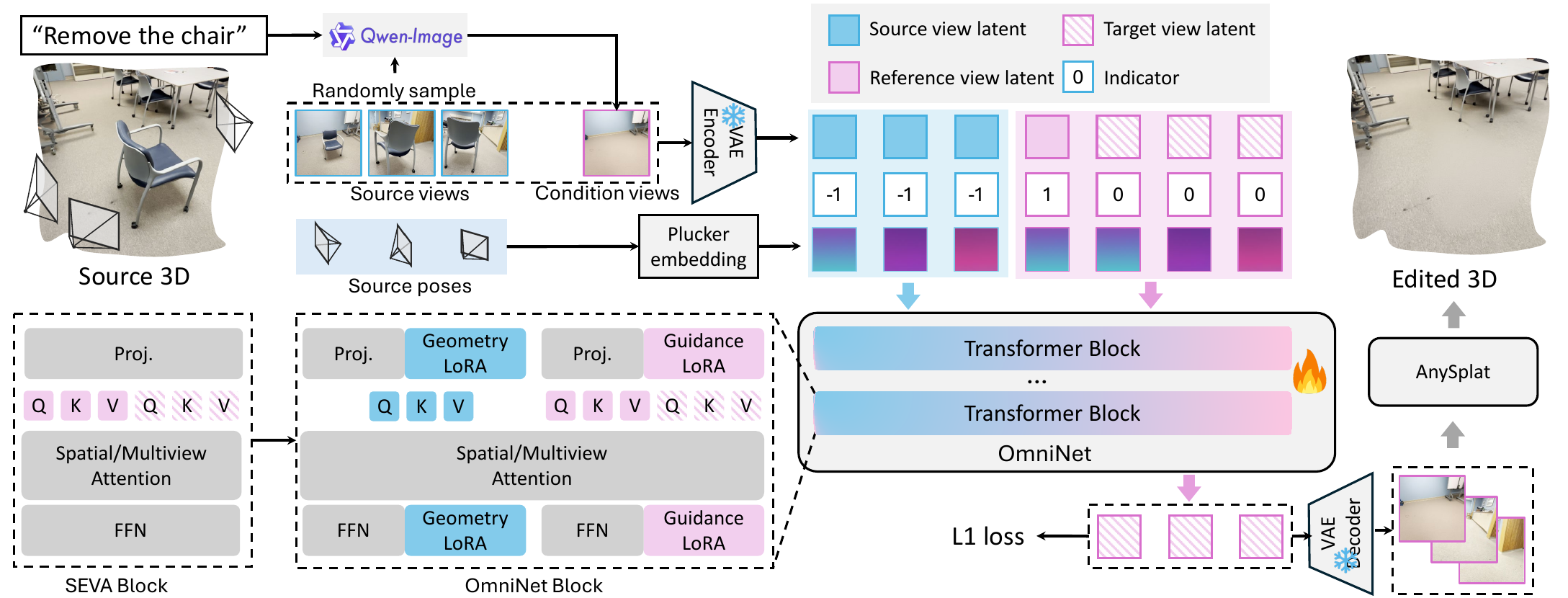}
        \caption{\textbf{Overview of Omni-3DEdit.} Given the instruction and multi-view images as inputs, we first employ Qwen-Image to obtain an edited reference image as condition view. Then an OmniNet is trained to  map the editing cues from condition view to other views. The outputs of OmniNet are edited multi-view images, which can be used to obtain the edited 3D asset optionally.}
    \label{fig:framework}
    \vspace{-3pt}
\end{figure*}

\noindent \textbf{3D Editing in Video/Multi-view Latent Space.}
Some methods use pre-trained video generation models~\cite{cogvideo, cogvideox, svd,wan22,yang2024direct, yang2026effectmaker} to predict edited multi-views within the RGB space across consecutive frames. The generative video models enable the handling of scene-level 3D editing.
DGE~\cite{dge}, EditCast3D~\cite{editcast3d}, and V2Edit~\cite{v2edit} introduce video editing for 3D editing.
However, video generation models suffer from (1) a weak prior for 3D consistency, (2) the need for continuous viewpoint transformations, which is computationally expensive, and (3) a lack of understanding camera poses, resulting in suboptimal performance in both computational efficiency and editing quality. 
DiGA3D~\cite{diga3d}, Pro3D-Editor~\cite{pro3deditor}, and methods~\cite{geometry, coupled, c3editor} share a similar idea of studying editing in multi-view latent space. They fall into the per-scene-optimization paradigm for specific editing tasks. Concurrent work Tinker~\cite{tinker} fails to tackle 3D editing that involves significant geometry changes, such as removal or addition.
Instead, our Omni-3DEdit achieves unified and generalized 3D editing.

\section{Method}

Our Omni-3DEdit firstly leverages the 2D multimodal editor Qwen-Image \cite{qwenimage} to perform instruction-based editing on an image from a randomly selected view, obtaining a \textit{edited reference image} as the condition. 
Then, an $\textit{OmniNet}$ is introduced and trained to propagate the edited cues to the other source views.
This paradigm not only takes advantage of the recent progress in 2D editing models but also reduces data and resource consumption, so that the OmniNet can focus exclusively on learning within the vision modality. 
In this section, we first describe the overview of Omni-3DEdit in Sec.~\ref{sec:method_overview} and then introduce the pipeline to generate paired training data in Sec.~\ref{sec:method_data}. The proposed dual-stream LoRA for training OmniNet is elaborated in Sec.~\ref{sec:dual_stream_lora}.

\subsection{Overview of Omni-3DEdit}
\label{sec:method_overview}

As shown in Fig.~\ref{fig:framework}, given $N$ source view inputs  $I_{src}=\{I_{src}^1, I_{src}^2,...,I_{src}^N\}$ from the source 3D scene and the editing instruction prompt $P$, we first employ VGGT~\cite{vggt} to obtain their relative camera poses $p=\{p^1, p^2, ..., p^N\}$. Then, we randomly select a view from the input views and edit it using off-the-shelf Qwen-Image~\cite{qwenimage}, obtaining a conditional image $I_{cond}$ to provide editing cues. These images are fed into a VAE encoder~\cite{sd} to produce the source view latents $s=\{s^1, s^2, ...,s^N\}$ and the condition view latent $c$.

During the training phase,  noisy target latents $y_{\sigma}=\{y_{\sigma}^1, y_{\sigma}^2, ...,y_{\sigma}^N\}$ are obtained following EDM~\cite{edm}:
\begin{equation}
    y_{\sigma}^n = y^n + \sigma \epsilon, 
\end{equation}
where $y^n$, $\sigma$, and $\epsilon$ are the $n$-th clear target view latent, noise level, and random noise, respectively. 
We concatenate the triplet latents $s$, $c$, and $y_{\sigma}$ in sequence space to avoid introducing an extra module, taking full advantage of the pretrained prior for understanding geometry relations among views.
To distinguish the different latents, $s$, $c$, and $y_{\sigma}$ receive -1, 1, and 0 indicators in feature space, respectively. 
Besides, to supplement the perspective geometry relations among views, $p$ of source views are converted into Plücker embeddings and conveyed to the condition view and noisy target views in the feature space. 
These input cues are fed into OmniNet $f(\cdot)$ to perform sample prediction.
Similar to the training paradigm of SEVA~\cite{seva}, the loss is calculated only for the latents of target views, as shown below:
\begin{equation}
    \mathcal{L} = \mathbb{E} \left[ \| (f(y_{\sigma}, s, c, \sigma) - y \|_2^2 \right].
\end{equation}

During the inference stage, the edited views can be interpolated by feeding the denoised target latent into VAE decoder. AnySplat~\cite{anysplat} is optional to obtain edited 3D assets based on edited multi-view in seconds.

{Note that Omni-3DEdit makes no assumptions about task priors. Instead, the model needs to implicitly learn to propagate the edit content solely based on the relationship between the reference and source views, thereby ensuring compatibility with versatile tasks. 
}

\begin{figure*}[ht!]
    \centering
    \includegraphics[width=1\linewidth]{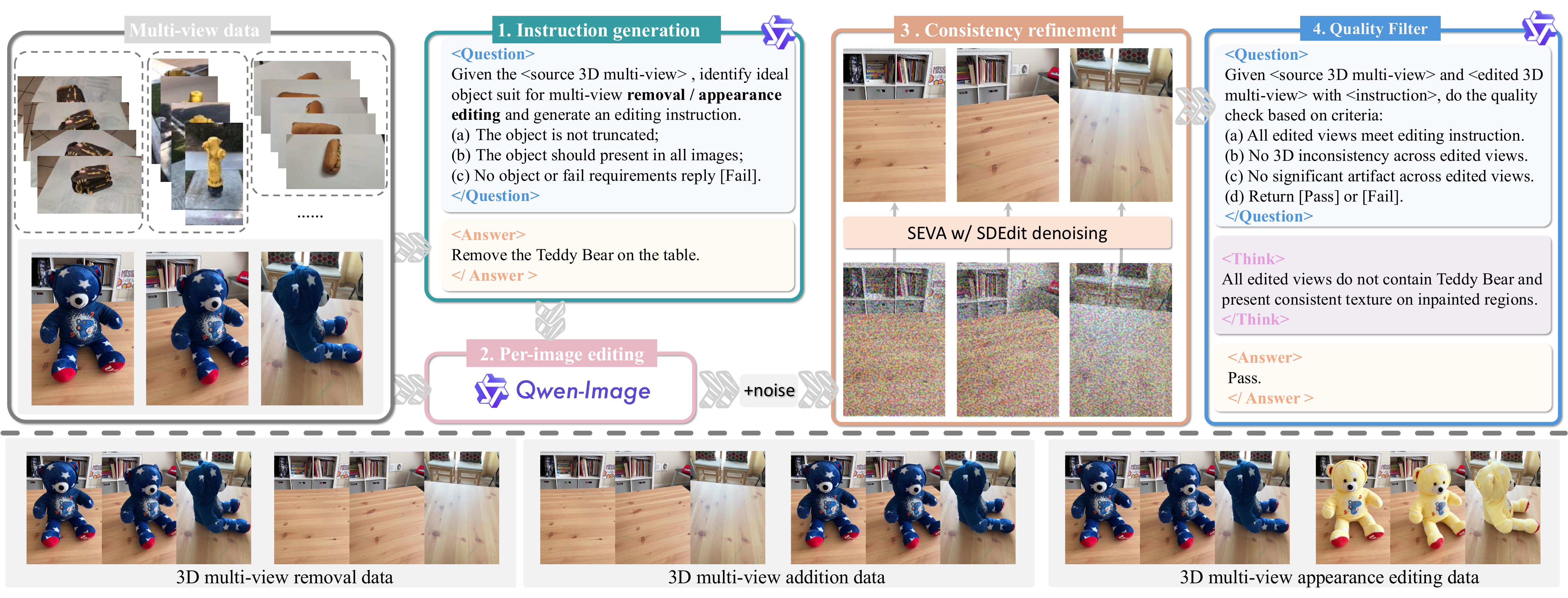}
    \vspace{-4mm}
        \caption{\textbf{Data Construction Pipeline.} The original multi-view images are passed through a four-stage pipeline to obtain their paired multi-view counterparts after editing. The pipeline covers tasks of 3D removal, addition, and appearance editing.}
    \label{fig:data_pipeline}
\end{figure*}

\subsection{Paired Training Data Generation}
\label{sec:method_data}
To drive OmniNet training, we require a dataset of paired 3D multi-view images before and after editing, which should cover diverse scenes and various editing tasks. Given the scarcity of publicly available large-scale scene-level editing datasets, we focus on three common editing categories: 3D addition, 3D removal, and appearance editing. Leveraging existing open 3D multi-view datasets, we establish a data pipeline that integrates existing tools to batch-generate the desired samples for training.
{Our key insight lies in the observation that both per-view 3D removal and appearance editing usually involve slight multi-view texture inconsistency, which can be alleviated via consistent refinement, while 3D addition data can be obtained by inverting the source and edited views of 3D removal data.}

\noindent \textbf{3D Removal.} The pipeline is illustrated in Fig.~\ref{fig:data_pipeline}, which includes 4 steps. (1) \textit{Instruction gneration.} Given multiple source views, we first employ Gemini-2.5pro~\cite{gemini} to analyze and identify an ideal object for deletion, and concurrently generate a textual editing instruction. Suitable candidates are defined as objects that have clear boundaries, are not truncated, and remain consistently visible across all views. (2) \textit{Per-image editing.} We utilize  Qwen-Image~\cite{qwenimage} to perform per-view foreground removal with generated instructions. (3) \textit{Consistency refinement.} To alleviate inconsistencies introduced by per-view editing ($\eg$, object-removed background regions exhibit disparate textures or colors),  inspired by SDEdit~\cite{sdedit}, we introduce light-intensity (20\%) noise to all edited views and then denoise them using the pre-trained SEVA~\cite{seva}. (4) \textit{Quality filter.} Due to the inherent stochasticity of 2D editing models and the varying difficulty of editing certain objects, which can lead to undesirable results, we implement an additional quality assessment to filter out undesired or failed samples.

\noindent \textbf{3D Appearance Editing.} 
Similar to the 3D removal task, appearance editing presents challenges in maintaining visual consistency across views when they are edited frame-by-frame. These inconsistencies manifest with variations in texture and color. Fortunately, such issues can be effectively addressed through the consistency refinement step in our 3D removal data pipeline.

\noindent \textbf{3D Addition.} 
In contrast to 3D removal and appearance editing, 3D addition introduces severe geometric inconsistencies across the edited views, which cannot be effectively alleviated through consistency refinement. Therefore, we adopt a reverse strategy for data generation. Specifically, we utilize the original 3D multi-view images as the target views, and subsequently use the outputs from our 3D removal data pipeline as the corresponding source views. This methodology ensures that the ground-truth target views are inherently multi-view consistent. This strategy is similar to VIVID-10M~\cite{vivid10m}, but we do not need extra masks~\cite{zhang2025bevdilation,zhang2024general,chen2023fpr,chen2020weakly}.

\begin{table}[t!]
\centering
\caption{{\textbf{Statistics of paired edited multi-views} in our curated training set, categorized by dataset and edit type.}}
\vspace{-1mm}
\resizebox{1\linewidth}{!}{

\begin{tabular}{l| ccc} 
\toprule
Editing Type & CO3Dv2~\cite{co3d} & DL3DV~\cite{dl3dv} & WildRGB-D~\cite{wildrgbd} \\
\midrule
3D Addition & 15K & 7K & 6K \\  
3D Removal & 15K & 6K  & 6K \\ 
3D Appearance Editing & 13K & 5K & 5K \\  
\bottomrule
\end{tabular}
}
\vspace{-5mm}
\label{tab:statistic}
\end{table}

\begin{table*}[t!]
\centering
\caption{\textbf{Quantitative comparison (PSNR $\uparrow$ / LPIPS $\downarrow$) of $360^\circ$ 3D removal methods on the 360-USID dataset.}}
\label{tab:quantitative_comparison}
\vspace{-2mm}
\resizebox{\textwidth}{!}{%
\begin{tabular}{l | ccccccc|c} 
\toprule
Methods & Carton & Cone & Cookie & Newcone & Plant & Skateboard & Sunflower & Average \\
\midrule
SPIn-NeRF~\cite{spinnerf} & 16.659 / 0.539 & 15.438 / 0.389 & 11.879 / 0.521 & 17.131 / 0.519 & 16.850 / 0.401 & 15.645 / 0.675 & 23.538 / 0.206 & 16.734 / 0.464 \\
2DGS + LaMa~\cite{2dgs, lama} & 16.433 / 0.499 & 15.591 / 0.351 & 11.711 / 0.538 & 16.598 / 0.670 & 14.491 / 0.564 & 15.520 / 0.639 & 23.024 / 0.194 & 16.195 / 0.494 \\
2DGS + LeftRefill~\cite{2dgs,leftrefill} & 15.157 / 0.567 & 16.143 / 0.372 & 12.458 / 0.526 & 16.717 / 0.677 & 12.856 / 0.666 & 16.429 / 0.634 & 24.216 / 0.181 & 16.282 / 0.518 \\
LeftRefill~\cite{leftrefill} & 14.667 / 0.560 & 14.933 / 0.380 & 11.148 / 0.519 & 16.264 / 0.448 & 16.183 / 0.463 & 14.912 / 0.572 & 18.851 / 0.331 & 15.280 / 0.468 \\
Gaussian Grouping~\cite{gsgroup} & 16.695 / 0.502 & 14.549 / 0.366 & 11.564 / 0.731 & 16.745 / 0.533 & 16.175 / 0.440 & 16.002 / 0.577 & 20.787 / 0.209 & 16.074 / 0.480 \\
GScream~\cite{gscream} & 14.687 / 0.587 & 14.655 / 0.476 & 12.733 / 0.429 & 13.662 / 0.605 & 16.238 / 0.437 & 12.941 / 0.626 & 18.470 / 0.436 & 14.758 / 0.514 \\
Infusion~\cite{infusion} & 14.191 / 0.555 & 14.163 / 0.439 & 12.051 / 0.486 & 9.562 / 0.624 & 16.127 / 0.406 & 13.624 / 0.638 & 21.195 / 0.238 & 14.416 / 0.484 \\
Aurafusion360~\cite{aurafusion} & 17.675 / 0.473 & 15.626 / 0.332 & 12.841 / 0.434 & 17.536 / 0.426 & 18.001 / 0.322 &  17.007 / 0.559 &  24.943 / 0.173 &  17.661 / \textbf{0.388}\\
\hline
\textbf{Omni-3DEdit (Ours)} & 18.578 / 0.452 & 16.296 / 0.360 & 13.183 / 0.465 & 17.829 / 0.336 & 17.598 / 0.411 & 16.764 / 0.521 & 23.801 / 0.224 & \textbf{17.722} / 0.395 \\

\bottomrule
\end{tabular}%
} 
\label{tab:quantitative_360usid}
\vspace{-7pt}
\end{table*}

\noindent \textbf{Statistics of Constructed Training Pairs.} With the training data generation pipeline, 
We construct training data based on three off-the-shelf multi-view datasets: CO3Dv2~\cite{co3d}, DL3DV~\cite{dl3dv}, and WildRGB-D~\cite{wildrgbd}, to cover a diverse range of 3D scenes from both indoor and outdoor. 
For each dataset, we begin by uniformly sampling scenes across different categories. From each selected scene, we randomly sample 20 images as training views. 
The final number of paired images in our training set is shown in Tab.~\ref{tab:statistic}. {Note that due to the high complexity of the scenes in DL3DV and WildRGB-D, relatively fewer samples passed this quality filtering, resulting in fewer training pairs from these sources. During the training phase, we sample the editing tasks and scenes uniformly.
}

\subsection{Dual-stream LoRA}
\label{sec:dual_stream_lora}
With the constructed paired data, we repurpose SEVA~\cite{seva, wang2026one2scene} as our editing model OmniNet, and finetune it to achieve multi-view editing by receiving additional source view latents $s$ alongside the reference (condition) latent $c$ and target view latents $y_{\sigma}$.
To incorporate source view cues, there are two typical manners of feature-space concatenation and sequence-space concatenation, which have been adopted in previous novel view generation studies~\cite{viewcrafter, recammaster}.
However, we observe a significant performance degradation for these two architectures. As shown in Fig.~\ref{fig:ablation_arch}, repurposed SEVA not only loses generative capability in target regions but also fails to preserve the unedited context from the source view.


We attribute this phenomenon to the use of shared projection layer for processing functionally distinct inputs. First, the source views and condition view serve different purposes. The condition view provides a precise editing signal from a specific perspective, whereas the source view provides comprehensive original context and texture information across camera poses.
Forcing OmniNet to process these functionally dissimilar latents with shared layers introduces a learning conflict. Second, the shared weights lack the ability to differentiate bias signals across blocks, so that the model has to distinguish source and condition latents by mapping them to different feature spaces. This mechanism is ineffective in helping the target latent to correctly identify and utilize the source latent features.

Therefore, as shown in Fig.~\ref{fig:framework}, we modify the architecture of SEVA to maintain two distinct sets of parameters within each block to individually encode the source latent and the condition latent.
Specifically, OmniNet is based on the pre-trained linear layers of SEVA and introduces a dual-stream LoRA~\cite{lora} module, including a geometry LoRA to process $s$ to capture geometry priors among source views, and a guidance LoRA to propagate editing guidance from $c$ to $y_{\sigma}$. The features from dual streams will exchange geometry cues and editing guidance in shared multi-view attention layers.
This distengled mechanism not only enables OmniNet to learn specialized representations from different views but also introduces a crucial inductive bias, which ensures that the target latent can correctly identify and attend to the features from both view latents.

\noindent \textbf{Discussion.} Note that compared to MM-DiT~\cite{mmdit}, our proposed dual-stream LoRA has two notable distinctions. First, MM-DiT maintains two independent sets of full parameters, but our method utilizes parameter-efficient LoRA modules. This allows OmniNet to leverage the priors of SEVA without full-scale duplication. Second, MM-DiT is designed to handle inputs from different modalities (\eg, text and image). In contrast, we prove that such a dual-stream paradigm is  effective for inputs of the same modality (\ie, vision latents) that serve distinct roles.

\begin{figure*}[ht!]
    \centering
    \includegraphics[width=1.\linewidth]{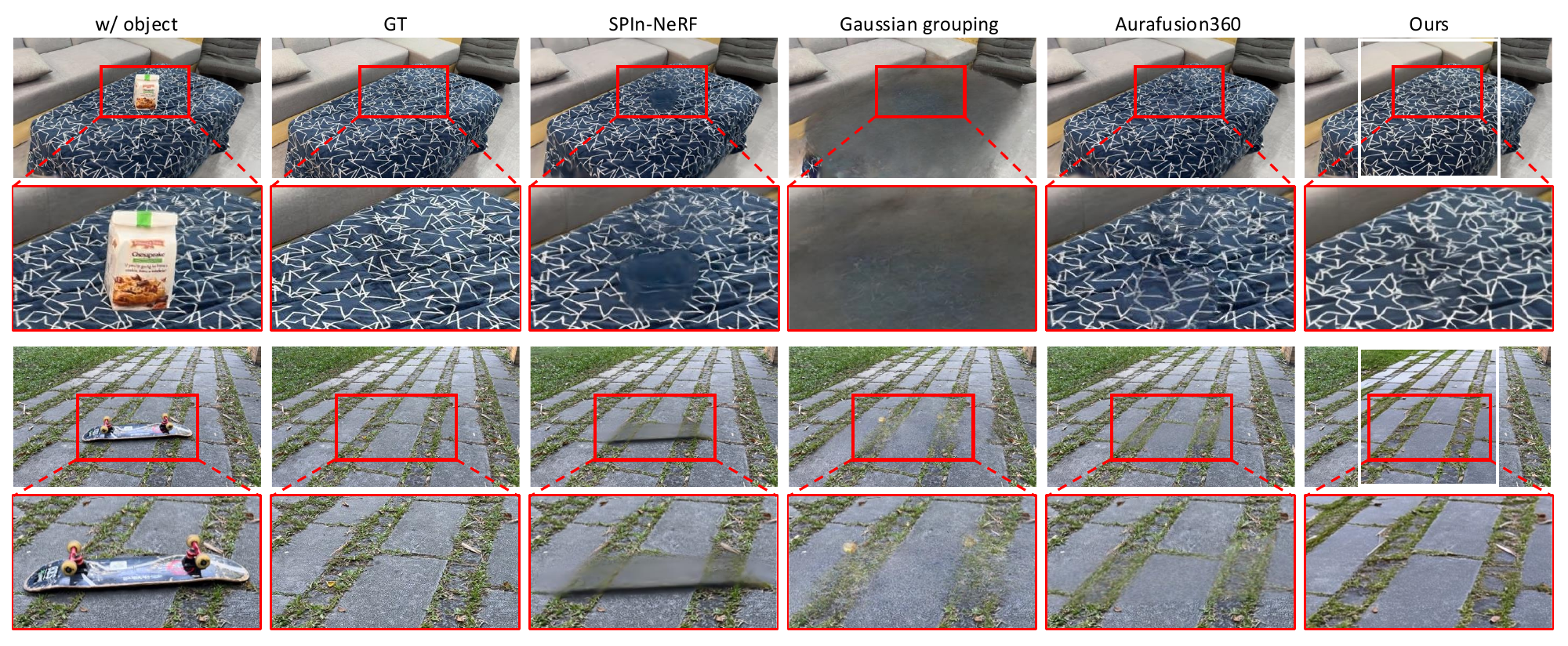}
    \vspace{-6mm}
    \caption{\textbf{Qualitative comparisons to 3D removal methods.} Our Omni-3DEdit not only removes the specific object completely but also presents rich details in the removed regions compared to other methods. We center crop views to adapt OmniNet resolution (white box).}
    \label{fig:360usid}
    \vspace{-10pt}
\end{figure*}

\section{Experiment}

\subsection{Experiment Setup}
\noindent \textbf{Implementation Details.}
For preprocessing, we follow the pipeline from SEVA~\cite{seva}, normalizing cameras within the same scene to the coordinate range of $[-2, 2]$ and setting $N=10$. The LoRA rank is set to 8. We train OmniNet for 4,000 iterations with batch size of 32, distributed across 16 NVIDIA H20 GPUs. The number of multi-view denoising steps is set to 50. All images are processed at a resolution of $576 \times 576$. We utilize the AdamW~\cite{adamw} optimizer with a constant learning rate of $1 \times 10^{-4}$ with Eps-weighting MSE loss. We follow SEVA~\cite{seva} to use SNR shift.

\noindent \textbf{Evaluation Datasets.}
While Omni-3Dedit can tackle Versatile tasks in an unified manner, we compare it with specific methods respectively.
For 3D removal, we utilize the unbounded 360 scene dataset, 360-USID~\cite{aurafusion}, 
which contains seven 360-degree scenes, comprising three indoor and four outdoor environments. 
For 3D addition, we follow MVInpainter~\cite{mvinpainter}
to study the NVS performance on CO3Dv2~\cite{co3d} validation set with sampling one scene per object. 
For 3D appearance editing, we omit numerical results due to the lack of publicly available benchmarks. Instead, we collect a series of complex 3D editing cases involving multi-round editing or significant geometry changes to compare the performance of different methods.

\noindent \textbf{Metrics.}
We use PSNR, and LPIPS as our evaluation metrics on 3D removal and addition. Following the protocol in prior work~\cite{spinnerf}, we compute these metrics only within the object mask to more accurately evaluate the result of 3D removal. Besides, we introduce CLIP text-image score, CLIP directional score to study the editing quality on our curated test set. In addition, we leverage mLLM Gemini-2.5pro~\cite{gemini} to conduct a more comprehensive evaluation on the 3D editing quality. Please refer to \textbf{Supplementary File} for more details. 

\subsection{Experimental Results}

\textbf{3D Removal}.
Omni-3DEdit is directly applicable to the 3D removal task. By using a 2D editor to perform object erasure on an arbitrary single view, we acquire a reference anchor view. OmniNet can then generate all remaining views. Note that our method operates in a mask-free manner, in contrast to prior works such as MVinpainter~\cite{mvinpainter}, SPINNeRF~\cite{spinnerf}, and Aurafusion360~\cite{aurafusion}, which need multi-view object masks to localize the target regions.

We first conduct a quantitative evaluation on the 360-USID dataset~\cite{aurafusion}, comparing our method against specialized 3D removal baselines, including 2DGS~\cite{2dgs} + LeftRefill~\cite{leftrefill}, GScream~\cite{gscream}, and SPIn-NeRF~\cite{spinnerf}. As demonstrated in Tab.~\ref{tab:quantitative_360usid}, our method achieves superior 3D removal performance. Compared to Aurafusion360~\cite{aurafusion}, our approach achieves an advantage in PSNR and costs much less time (2min $v.s.$ 30min) since Omni-3DEdit is free of iterative warping and obtains edited multi-view in a single pass.

We then provide qualitative comparisons in Fig.~\ref{fig:360usid} to illustrate the superiority of our approach. One can observe that compared to Gaussian Grouping~\cite{gsgroup}, Omni-3DEdit correctly identifies the target object for removal without corrupting the content of adjacent objects (\eg, the desk). Furthermore, regarding the visual quality of object-removed regions, Aurafusion360~\cite{aurafusion} exhibits significant artifacts and residual contours at the object boundaries, while our method demonstrates a clear advantage in maintaining high-fidelity and consistent details.

\begin{table}[t!]
    \centering
    \caption{\textbf{Quantitative comparison on CO3Dv2 \textit{val} set.} }
    \label{tab:quantitative_comparison_co3d}
    \vspace{-3mm}
    \resizebox{0.9\linewidth}{!}{
        \begin{tabular}{l|cccc}
            \toprule
            Method  & PSNR $\uparrow$ & LPIPS $\downarrow$ & CLIP-T $\uparrow$  \\
            \midrule
            ZeroNVS~\cite{zeronvs} & 14.56 & 0.716 & 0.196 \\
            MVInpainter~\cite{mvinpainter} & 19.20 & 0.344 & 0.271 \\
            \textbf{Omni-3DEdit (Ours)} & \textbf{20.67} & \textbf{0.278} & \textbf{0.277} \\
            \bottomrule
        \end{tabular}
    }
    \label{tab:nvs_co3d}
    \vspace{-3mm}
\end{table}

\vspace{+1mm}
\noindent\textbf{3D Addition}.
To validate the model's capability for 3D object addition, we follow the evaluation methodology established by MVInpainter~\cite{mvinpainter}, utilizing the multi-view images from the CO3Dv2~\cite{co3d} validation set. For each scene, we retain an arbitrary view as the reference image, while the foreground objects in all remaining views are erased and inpainted via Qwen-Image. 
OmniNet is employed to generate these erased objects conditioned on the reference image, where the target object visible. This experimental setup is to investigate the novel view synthesis (NVS) capability in the context of object addition.
As shown in Tab.~\ref{tab:nvs_co3d}, 
ZeroNVS~\cite{zeronvs} fails to fully take the context from source views and generates target views based on the single reference view, achieving the worst performance.
MVInpainter~\cite{mvinpainter} heavily relies on complex pre-processing (\eg, point matching, mask propagation), limiting its generalization ability and achieving sub-optimal performance. Our Omni-3DEdit learns the mapping from source views to target edited views and outperforms MVInpainter in synthesized novel view quality. Omni-3DEdit is built upon SEVA, thereby inheriting its original ability for consistent generation under specified camera poses.

\begin{figure}[t!]
    \centering
    \includegraphics[width=1.\linewidth]{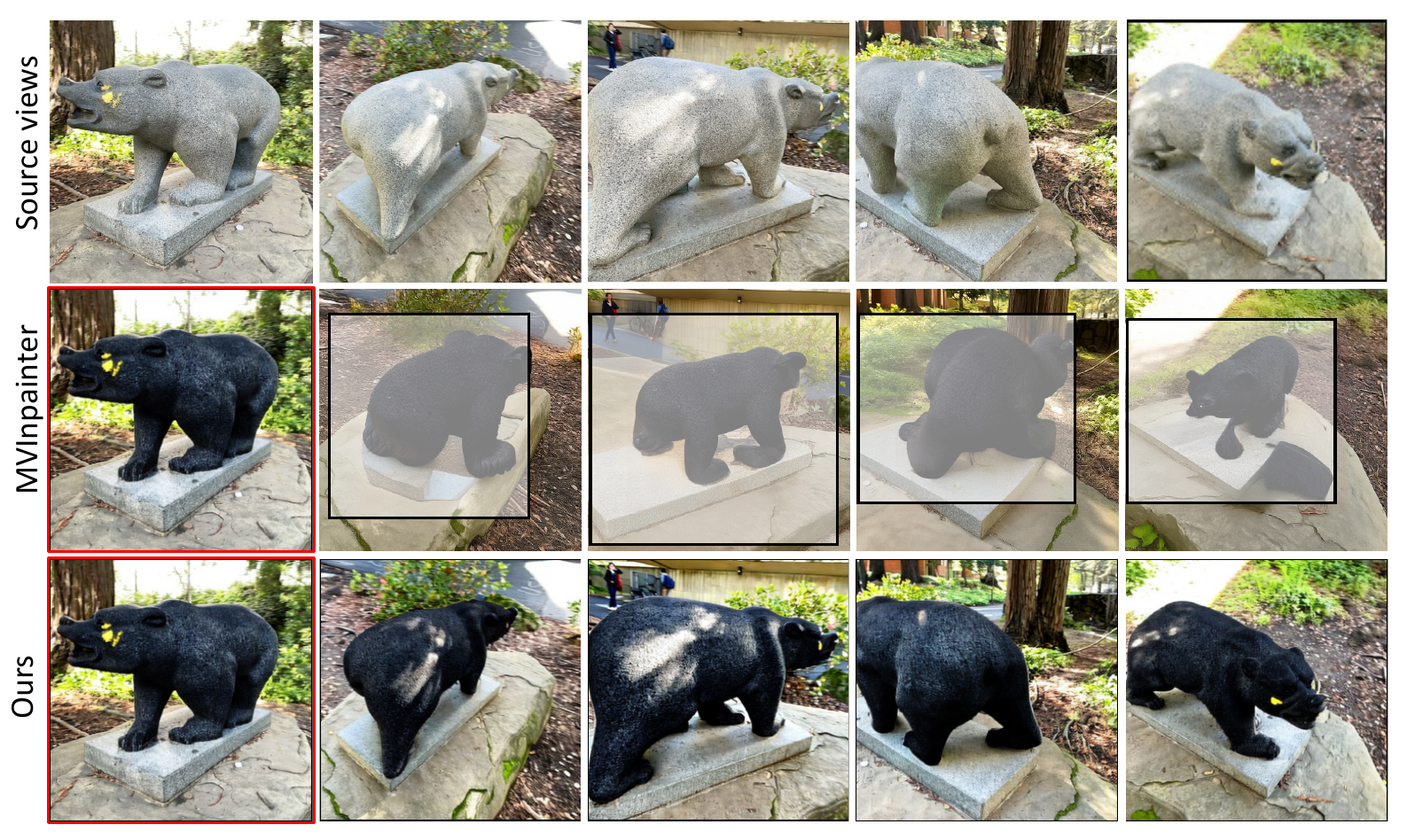}
    \vspace{-20pt}
    \caption{\textbf{Comparison of 3D appearance editing.} Red box represents the reference view. White boxes are object masks as additional inputs for MVInpainter.}
    \label{fig:bear_black}
    \vspace{-12pt}
\end{figure}

\vspace{+1mm}
\noindent\textbf{3D Appearance Editing}.
We further demonstrate the applicability of our method to 3D appearance editing. Despite relying on a single reference image that often offers limited-view guidance, Omni-3DEdit implicitly captures instance-level geometric priors, allowing it to effectively propagate editing signals to other regions unobserved in the reference view. 
We illustrate this capability with showcase of \textit{``Make bear black."}~\cite{instructn2n}, a scene out of our training data. 
As shown in Fig.~\ref{fig:bear_black}, while the reference image solely presents the left perspective, Omni-3DEdit successfully propagates the editing guidance to the entire bear instance, including its rear view.
In contrast, prior reference-based 3D editing methods often struggle with 360° appearance editing. This limitation stems from their either heavy reliance on depth warping~\cite{moge}, which fails to handle accumulated errors across large viewpoint changes~\cite{see3d,splatflow}, or their need for explicit masks for instance identification, resulting in the corruption of original geometric information~\cite{mvinpainter,see3d}. As demonstrated in the second row, MVinpainter fails to preserve the original geometry, yielding unsatisfactory results.

\vspace{+1mm}
\noindent\textbf{Complex 3D Editing}.
Our method supports fundamental operations, including 3D removal, 3D addition, and appearance editing. By combining them, we can achieve more complex editing tasks such as replacement and multi-turn editing. We collect a test benchmark to study the performance of Omni-3DEdit on such tasks. More details can be found in the \textbf{Supplementary File}. 

As shown in Tab.~\ref{tab:complex_edit}, Omni-3DEdit presents significant advantages over previous methods in terms of both Gemini score and time cost. This is mainly due to fact that previous studies can only tackle specific editing tasks such as appearance editing. In addition, they heavily rely on iteratively evoking the 2D editor and explicit 3D representations, resulting in long convergence.

\begin{figure}[t!]
    \centering
    \includegraphics[width=1.\linewidth]{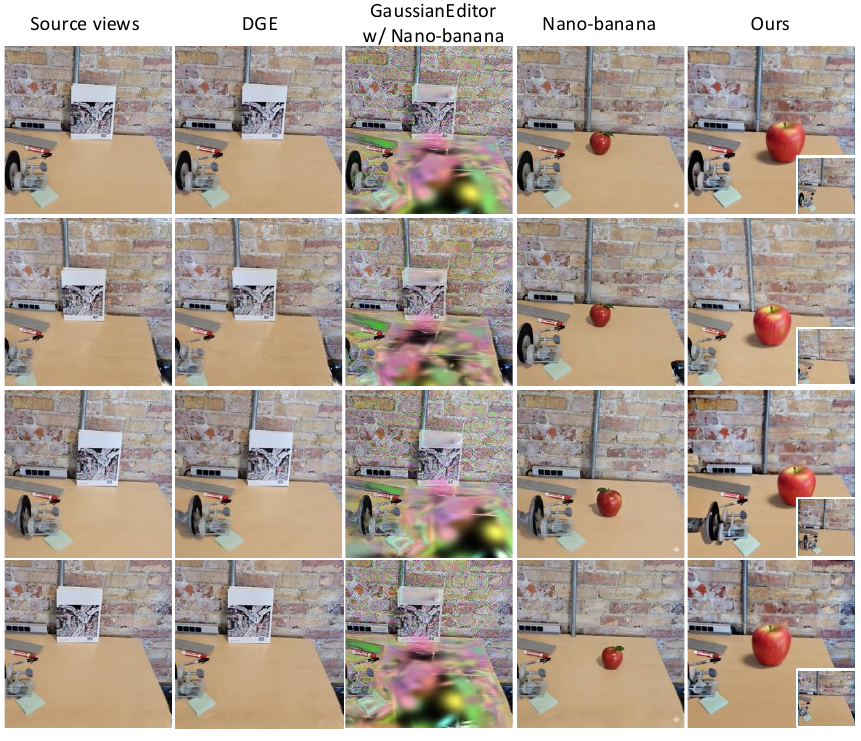}
    \vspace{-19pt}
    \caption{\textbf{Results of competing 3D editing methods for a complex task \textit{``Removing the book." then ``Adding an apple to desk.''}} }
    \label{fig:demo}
\vspace{-8pt}
\end{figure}

We present a showcase of \textit{``Removing the book."} then \textit{``Adding an apple to desk."} in Fig.~\ref{fig:demo}.
DGE~\cite{dge} fails to achieve clear 3D editing because it relies on the source geometry to find pixel correspondences. Similar failures occur for GaussianEditor~\cite{gaussianeditor}, although it is equipped with the more powerful editor, Nano-banana~\cite{gemini}. Since each edit might place the apple in a different position on the table, this makes it difficult for the explicit 3D Gaussian to converge, resulting in obvious artifacts.
In Nano-banana, we concatenate the source views into a single image, utilizing its in-context capability~\cite{incontext_edit} to edit multi-view at once, thereby improving their 3D consistency. However, the apples in edited views still suffer from inconsistent scale and position. 
In comparison, Omni-3DEdit maintains high consistency throughout this two-stage editing (removal results are shown in the right-bottom of last column), as evidenced by the coherence of wall tile textures and apple details.

\begin{table}[t!]
\centering
\caption{\textbf{Comparison of methods on complex 3D editing.}}
\vspace{-2mm}
\label{tab:complex_edit}
\resizebox{\linewidth}{!}{
\begin{tabular}{l|cccc}
\toprule
Method         & CLIP-T/I & CLIP-Dir. & Gemini score & Time \\
\midrule
DGE~\cite{dge}            &  0.246     & 0.132              &    1.7         &   5min   \\
GaussianEditor~\cite{gaussianeditor} &    0.253 &    0.146    &    2.0      & 17min     \\
ViCANeRF~\cite{vicanerf} & 0.257 & 0.141 &   2.2 & 28min \\
Nano-banana~\cite{gemini} & 0.281 & 0.165 & 3.8  & - \\
\hline
\textbf{Omni-3DEdit (Ours)}  & \textbf{0.286}     &  \textbf{ 0.170}              &   \textbf{4.0}          &  \textbf{2min}    \\
\bottomrule
\end{tabular}
\vspace{-17pt}
}
\end{table}

\begin{figure*}[ht!]
    \centering
    \includegraphics[width=1.\linewidth]{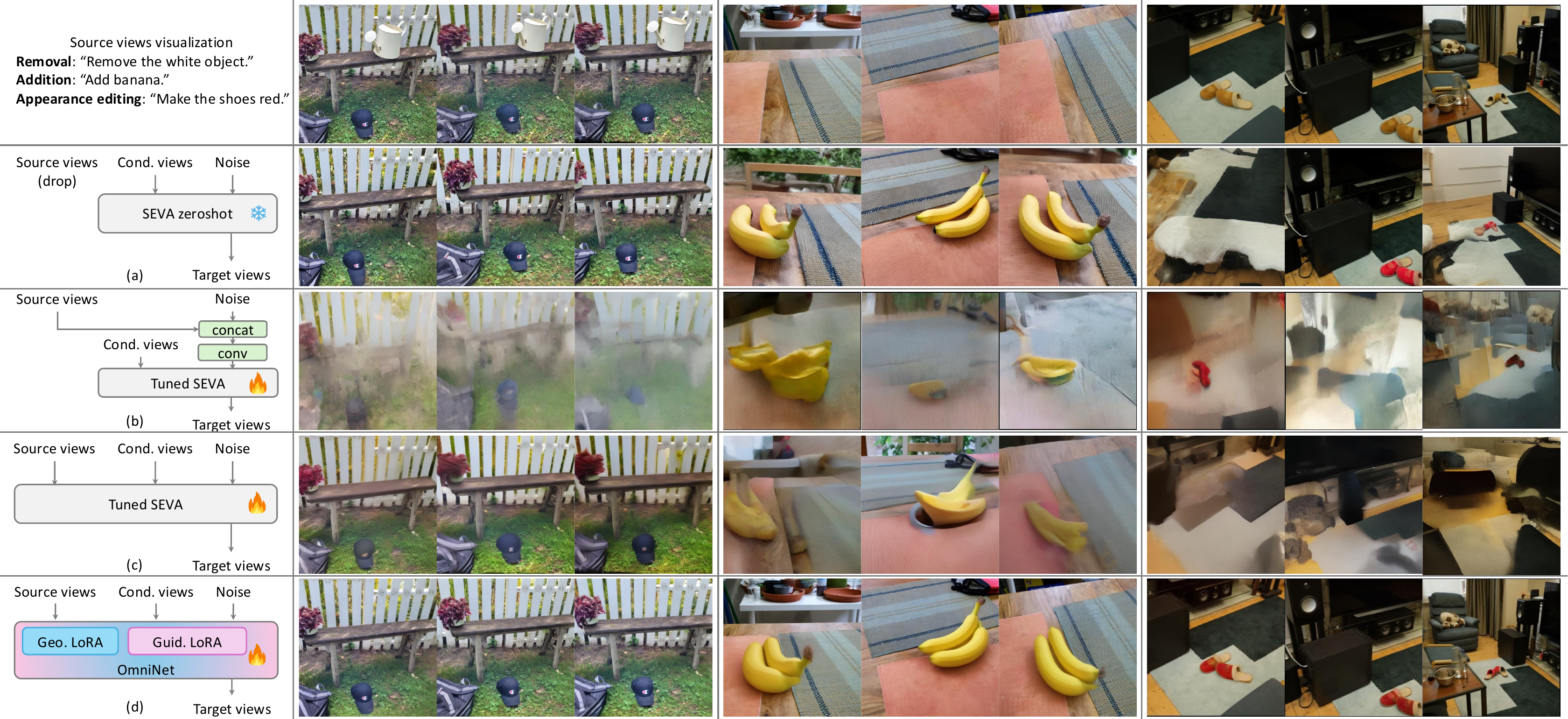}
    \caption{\textbf{Ablation study of different architectures on performing multiview editing with edited reference view.} Both plain feature-space concatenation and sequence-space concatenation fail to achieve desired editing results, while dual-stream LoRA improves the editing quality significantly. We provide edited multi-view here, and the edited 3D assets can be found in the \textbf{Supplementary File}.}
    \label{fig:ablation_arch}
    \vspace{-0pt}
\end{figure*}

\subsection{Ablation Study}
\noindent \textbf{Architecture.} We first investigate the impact of architectural choices on model performance by comparing our OmniNet, which is builtg upon sequence concatenation with dual-stream LoRA, with three distinct approaches that incorporate original images. 
(a) \textit{SEVA zeroshot}: dropping the source view and only feeding the reference view and Gaussian noises to pre-trained SEVA to obtain generated views under given camera poses.
(b) \textit{Feature-space concatenation}~\cite{viewcrafter}: concatenating each Gaussian noise with its corresponding source view latent along the channel dimension, followed by feature fusion via a lightweight projection network. (c) \textit{Sequence-space concatenation}~\cite{recammaster,fulldit}: concatenating the source view latent, condition view latent, and Gaussian noises along the sequence dimension, they share the same linear layers in each block. 

We provide visualization comparisons over three editing showcases provided in Fig.~\ref{fig:ablation_arch}. One can observe that SEVA zero-shot fails to align source view poses since the reliance on single conditional view and normalized camera poses makes SEVA scale-agnostic during the generation process. Feature-space concatenation tends to produce obvious artifacts with blurred details. We suspect it is caused by the fact that the light convolution layers are difficult to fuse cues from source views and target views, and the cues from source view latents are vanishing when passing through the network. 
A similar phenomenon of bypassing source view information is also observed in sequence-space concatenation. This reveals the difficulty for pre-trained parameters to simultaneously encode both source and target views, which provide distinct information.
By introducing dual-stream LoRA, OmniNet brings performance improvement by capturing geometry cues and editing guidance from source and condition views via decoupled LoRA parameters.

\vspace{+1mm}
\noindent \textbf{Input Signal}. Furthermore, we investigate the critical importance of input signals for OmniNet. We conduct ablation studies by dropping the indicator and camera pose to observe their respective impacts on model performance. Experiments on the 360-USID benchmark~\cite{aurafusion}, as presented in Tab.~\ref{tab:ablation_omninet}, reveal that performance degrades drastically without the indicator due to the lack of explicit signals distinguishing different views. Similarly, excluding camera poses leads to a significant performance drop, since relying solely on appearance to analyze perspective geometric information is too implicit for the model to effectively comprehend.

\begin{table}[t!]
  \centering
  \caption{\textbf{Ablation study on 360-USID benchmark.}}
  \label{tab:ablation_omninet}
  \vspace{-3mm}
  \resizebox{0.9\linewidth}{!}{
  \begin{tabular}{l|cccc}
    \toprule
    Settings & SSIM $\uparrow$ & PSNR $\uparrow$ & LPIPS $\downarrow$ \\
    \midrule
    Omni-3DEdit & 0.925 & 17.72 & 0.395 \\
    \hline
    SEVA zeroshot & 0.911 & 13.99 & 0.575 \\
    Omni-3DEdit w/o indicator & 0.917 & 15.20 & 0.545 \\
    Omni-3DEdit w/o pose & 0.903 & 14.54 & 0.565 \\
    \bottomrule
  \end{tabular}
  }
  \vspace{-3mm}
\end{table}

\section{Conclusion}
In this paper, we proposed Omni-3DEdit, a unified and generalized model capable of handling 3D removal, addition, and appearance editing without relying on additional masks or point matching signals. Specifically, we constructed high-quality paired edited multi-view data across different editing tasks and introduced a dual-stream LoRA module to repurpose the pre-trained multi-view generation model SEVA into a multi-view editing model, OmniNet. Extensive experiments on 3D removal, addition, and appearance editing tasks demonstrated that our Omni-3DEdit performs significantly better than existing schemes, showing strong generalization performance with much faster speed.

\vspace{+1mm}
\noindent \textbf{Limitations}.
Due to the scarcity of open-source scene-level multi-view data and computational resource constraints, the scale (0.1M) of our constructed dataset is relatively small, which limits Omni-3DEdit's ability to handle very fine-grained editing tasks (\eg, \textit{``adding a bracelet to a human wrist"}). A potential solution is to develop a more sophisticated data construction pipeline to generate a larger corpus of training data. In further, we plan to elaborate Omni-3DEdit with 3D agent~\cite{lin2025jarvisevo,lin2025jarvisart,zhang2025bevdilation,hu2025marketgen} or extend it into 4D paradigm~\cite{pan2025diff4splat,DynamicVerse}.

{
    \small
    \bibliographystyle{ieeenat_fullname}
    \bibliography{main}
}


\end{document}


\maketitle

\noindent This supplementary file provides the following materials:

\begin{itemize}
    \item  \textbf{Additional Training Data Details} (referring to Sec.~\ref{sec:training_data_detail});
    \item \textbf{Attention Visualization} (referring to Sec.~\ref{sec:add_exp});
    \item \textbf{Additional Testing Data Details} (referring to Sec.~\ref{sec:test_data_detail}).
\end{itemize}

\begin{figure}[h!]
    \centering
    \includegraphics[width=1.0\linewidth]{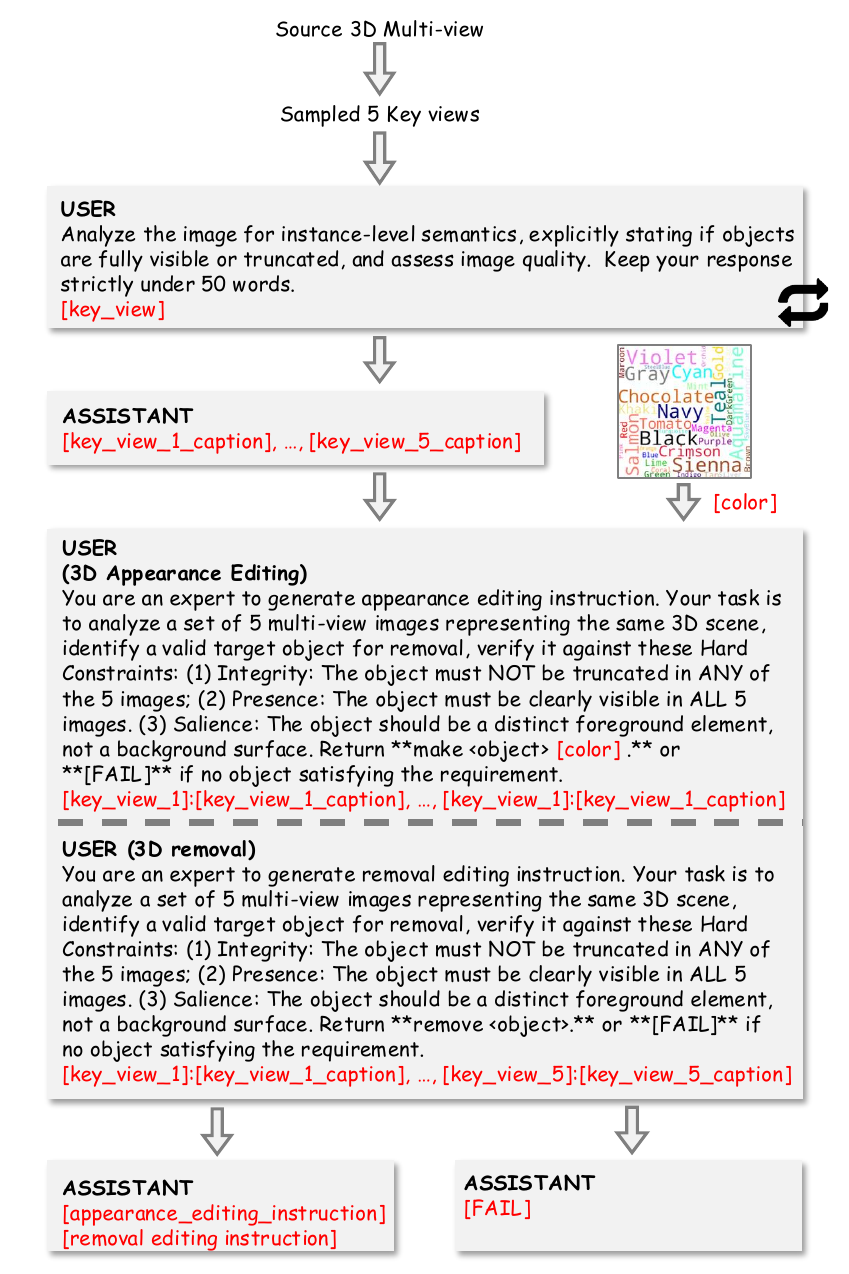}
    \caption{\textbf{System prompts for instruction generation.}}
    \vspace{-5pt}
    \label{fig:instruction_generation}
\end{figure}

\begin{figure}[t!]
    \centering
    \includegraphics[width=\linewidth]{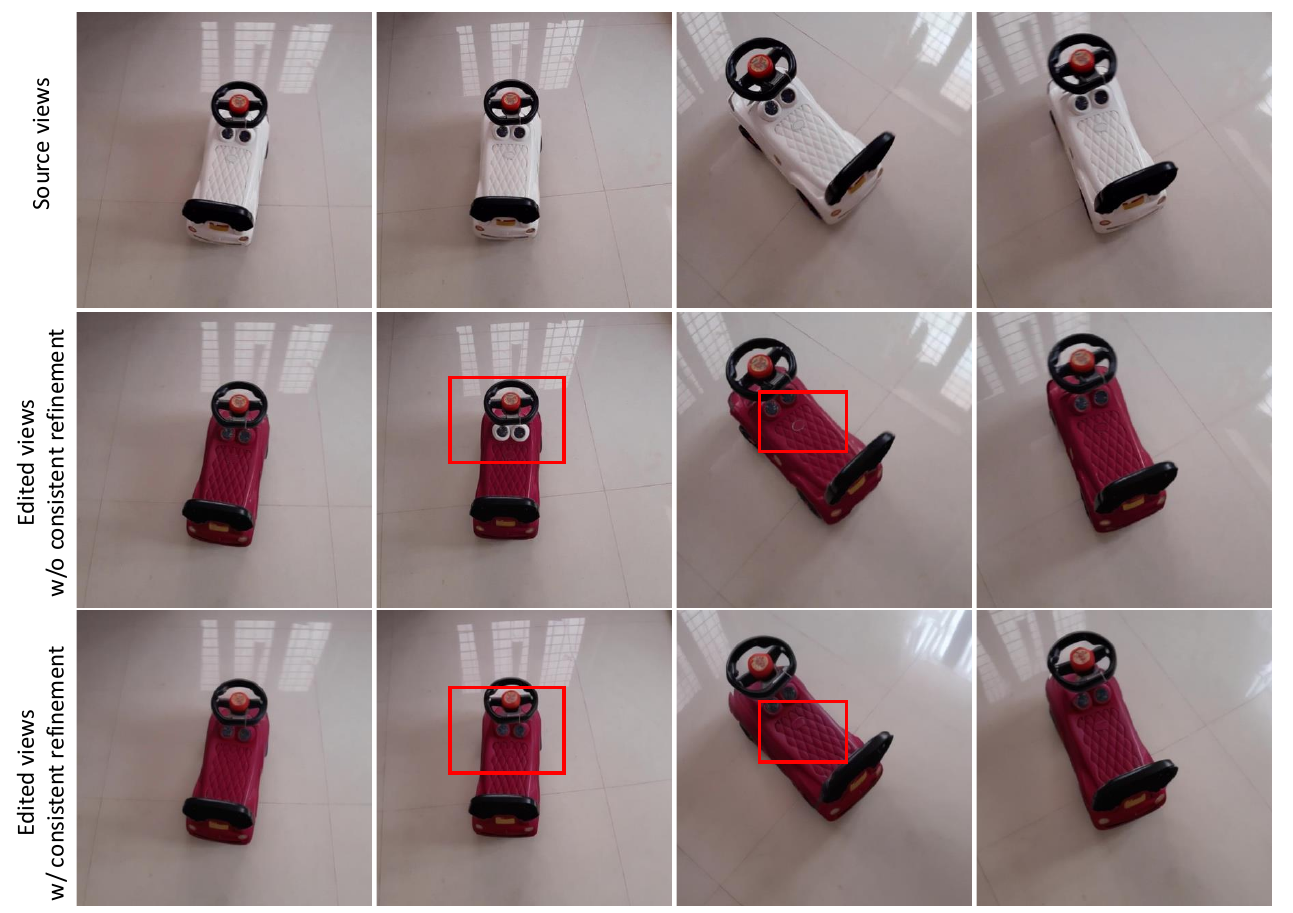}
    \caption{\textbf{Comparison of per-view editing and consistent refinement.} Consistent refinement eliminates the minor inconsistencies that occur on per-view appearance editing. }
    \label{fig:trainset_cr}
    \vspace{-7pt}
\end{figure}

\section{Additional Training Data Details}
\label{sec:training_data_detail}

In the main paper, we briefly described the four stages of our data curation pipeline. In this section, we provide more comprehensive implementation details.

\vspace{+1mm}
\noindent \textbf{Instruction Generation.} 
We employ a locally deployed Qwen2.5-VL-32B~\cite{qwen2} model together with the Gemini-2.5 Pro~\cite{gemini} API to generate appearance editing and object removal instructions based on the provided multi-view images. Given that processing an excessive number of views increases computational complexity and reduces inference speed, we randomly sample five views per scene as input for the model. 
Furthermore, to enhance multimodal reasoning capabilities, we first utilize VLM to describe the content of each image before determining the final editing instruction, which takes full advantage of the reasoning capability in the texture space. 
Regarding appearance editing, we observe that the model tends to generate repetitive colors ($e.g.$, frequently outputting ``black"), resulting in limited diversity. To address this, we establish a pre-defined color bank and specify a randomly sampled color for each appearance editing directly, significantly improving the diversity of the generated appearance editing instructions. The complete system prompts are provided in Fig.~\ref{fig:instruction_generation}.

\begin{figure*}[h!]
    \centering
    \includegraphics[width=\linewidth]{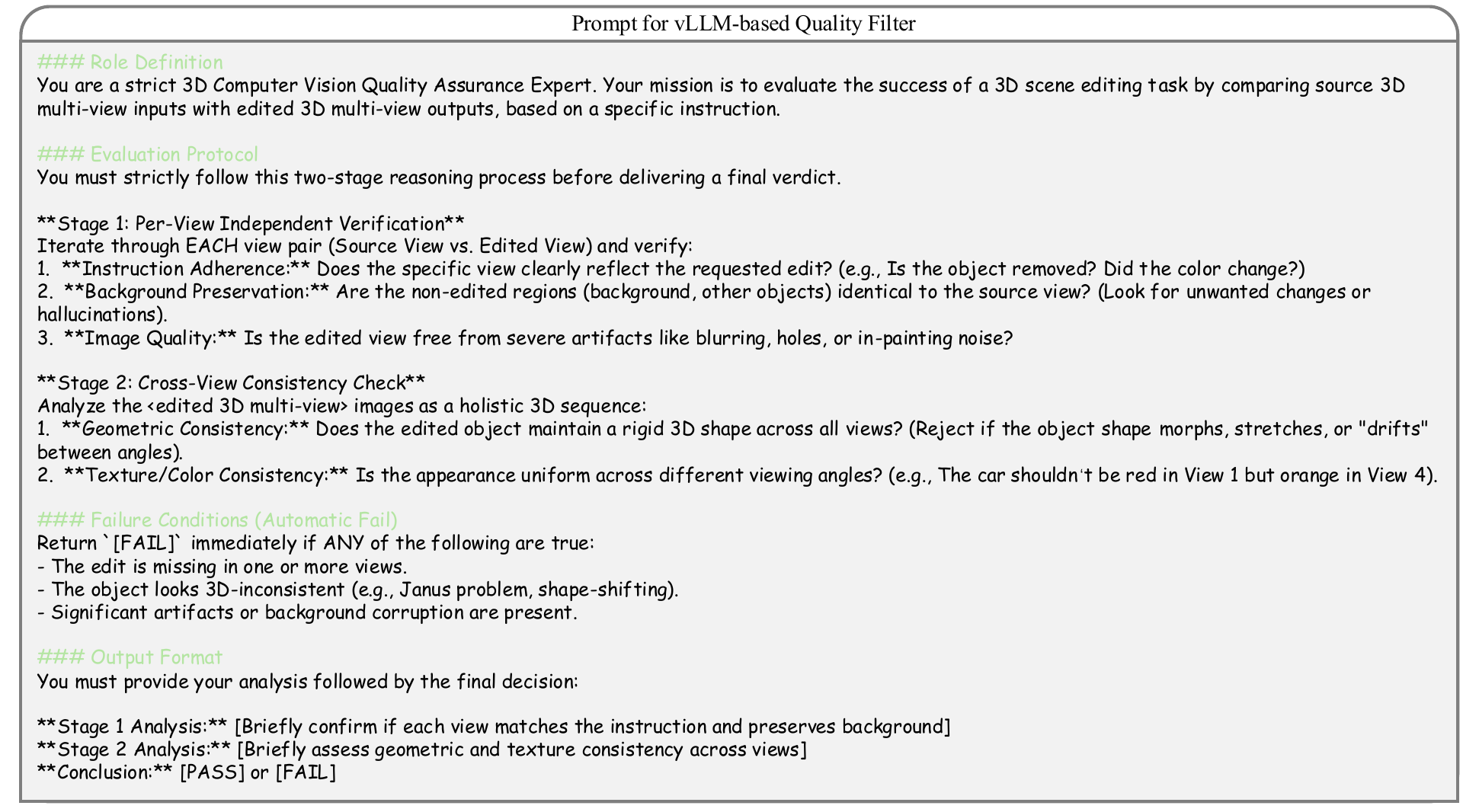}
    \caption{\textbf{System prompts for quality filter.} Firstly, evaluation is conducted on per source-edited views to ensure that the instruction has been successfully executed. Then, a subjectively multi-view consistent evaluation is conducted to filter cases with obvious inconsistency. }
    \label{fig:quality_filter}
\end{figure*}
\vspace{+1mm}

\noindent \textbf{Consistent Refinement.}
We observe that 3D appearance editing typically introduces minimal geometric deformation, while 3D object removal often results in backgrounds with smooth, regular geometry ($e.g.$, flat ground). Consequently, even with per-frame editing, both appearance editing and removal tasks will introduce minor 3D inconsistencies manifesting as color variations or high-frequency texture artifacts. These discrepancies can be mitigated through techniques such as iterative 2D-3D-2D refinement~\cite{dge,vicanerf,gaussianeditor,syncnoise}. 
However, we eschew multi-round iterative refinement during our data generation phase, as this process is prohibitively time-consuming ($\sim$10 minutes per scene) and hinders scalable data curation.
Instead, drawing inspiration from SDEdit~\cite{sdedit}, we randomly select an edited view to serve as the conditional view. For the remaining views, we add a small amount of noise and perform EDM denoising using the pre-trained SEVA model. This streamlined process requires $\sim$ 20 seconds per scene, significantly reducing the time overhead for data construction.
In the inference (denosing) stage, SEVA denoises the noisy target view latent $x_t$ with the EDM solver~\cite{edm} as follows\footnote{For simplicity, we omit the $2^{\text{nd}}$ order correction. The noise level increases with the step $t$, which is opposite to the original paper~\cite{edm}. }:
\begin{equation}
    x_t = x_{t+1}+\frac{\sigma_t-\sigma_{t+1}}{\sigma_{t+1}}(x_{t+1}-D_{\theta}(x_{t+1}, t+1; I)) ,
\end{equation}
where $\sigma_t$ is the scheduled noise level at step $t \in [0, T]$. $D_{\theta}(x_t,t;I)$ is the estimated $x_0$ from $x_t$ conditioned on the first frame $I$, which is defined as follows: 
\begin{equation}
    D_{\theta}(x_t, t; I) = x_{t}-(c_{skip}^{t}x_{t} + c_{out}^{t}F_{\theta}(c_{in}^{t}x_{t}, c_{noise}^{t}; I) ) ,
\label{eq:edm}
\end{equation}
where  $c_{skip}^{t}$, $c_{in}^{t}$, $c_{out}^{t}$, and $c_{noise}^{t}$ are coefficients of the noise schedule in EDM.

Fig.~\ref{fig:trainset_cr} visualizes an example to demonstrate the effectiveness of our consistency refinement method. During the process of converting a white car to a red car by invoking Qwen-Image \cite{qwenimage} on a per-view basis, local inconsistencies arise; for instance, the regions surrounding the two black circles in the second column are not turned red. Our consistency refinement successfully rectifies these discrepancies, yielding a highly consistent editing result.

\begin{figure*}[t!]
    \centering
    \includegraphics[width=0.9\linewidth]{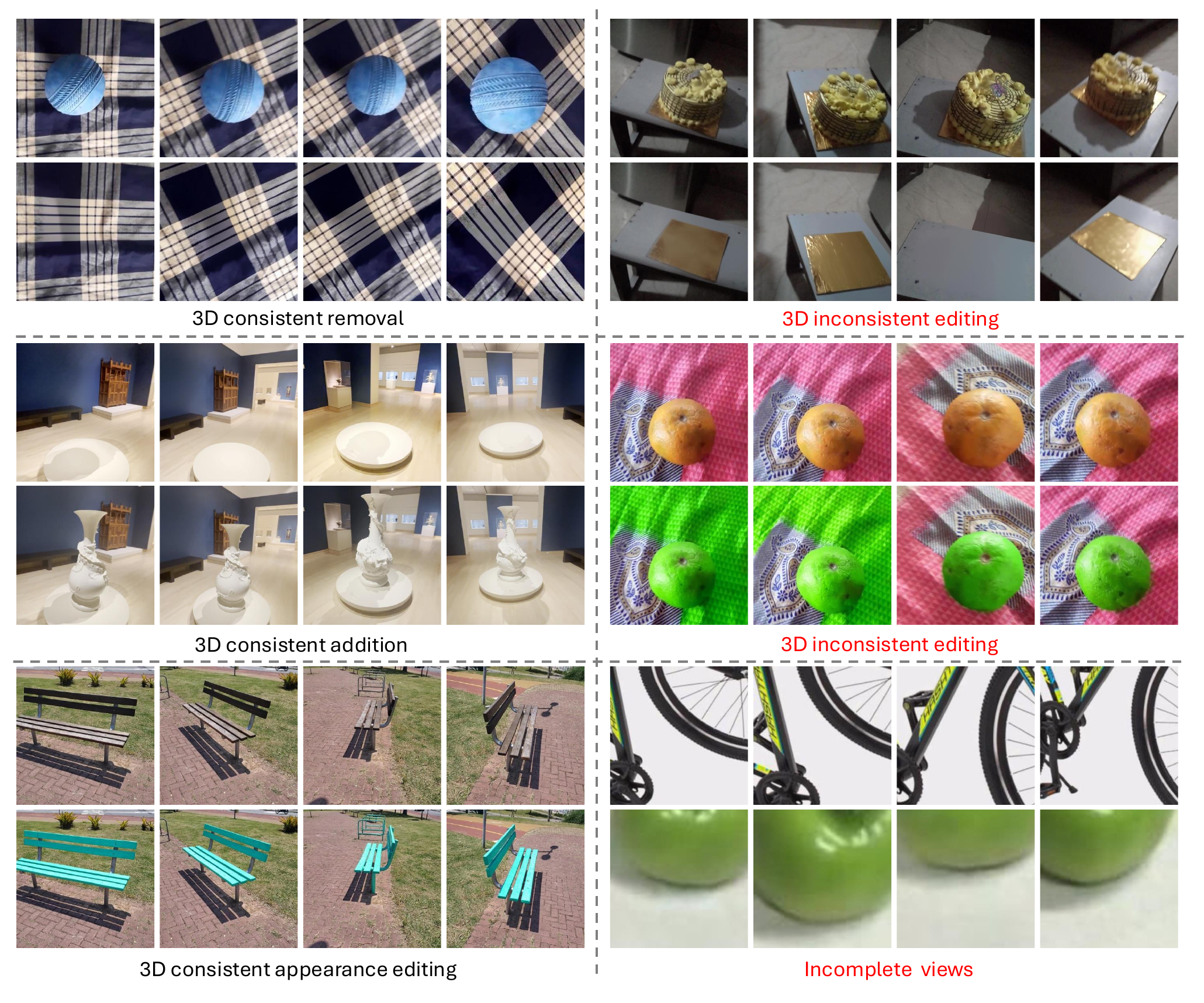}
    \caption{\textbf{Visualization of successful cases and failed cases.} We present several training data on the left and some failed cases on the right, which are caused by the inconsistent editing and incomplete views.}
    \label{fig:success_fail}
\end{figure*}

\vspace{+1mm}
\noindent \textbf{Quality Filter.}
Fig.~\ref{fig:quality_filter} provides a detailed illustration of the quality filtering process. We employ the Qwen2.5-VL-32B~\cite{qwen2} to verify whether the edited images align with the editing instructions relative to the original images, ensuring that irrelevant regions remain largely unchanged and the image quality is not degraded by editing. 
Subsequently, we employ the VLM to subjectively assess the multi-view consistency, filtering out samples with significant discrepancies that cannot be rectified through consistency refinement.

Fig.~\ref{fig:success_fail} visualizes some examples to demonstrate data curation results. On the left, we display successful cases that pass all the four curation stages, including the final quality filter. On the right, we show some failure cases. In case 1, the cardboard at the base of the cake was not correctly removed in certain views. In case 2, large areas of the table-cloth were erroneously colored green during the process of changing the orange to green. These two cases exhibit significant erroneous regions that could not be corrected by consistency refinement, and they are consequently dropped during quality filtering. Additionally, case 3 is also discarded during the instruction generation phase due to dataset limitations or cropping issues, where the views do not contain the complete object and cannot be recognized by VLM.

\begin{table}[t]
    \centering
    \caption{\textbf{Training settings and hyperparameters.}}
    \label{tab:training_settings}
    \resizebox{0.6\linewidth}{!}{%
        \begin{tabular}{l|l}
            \toprule
            \textbf{config} & \textbf{value} \\
            \midrule
            optimizer & AdamW \\
            base learning rate & 1e-4 \\
            weight decay & 0.01 \\
            eps & 1e-8 \\
            optimizer momentum & $\beta_1, \beta_2=0.9, 0.999$ \\
            batch size & 32 \\
            learning rate schedule & ConstantLR \\
            training steps & 4K \\
            loss weight & EpsWeight \\ 
            snr shift & 2.4 \\
            CFG & 1.2 \\
            UCG rate & 0.2 \\ 
            \bottomrule
        \end{tabular}%
    }
    \label{tab:}
\end{table}

\begin{figure*}[t!]
    \centering
    \includegraphics[width=\linewidth]{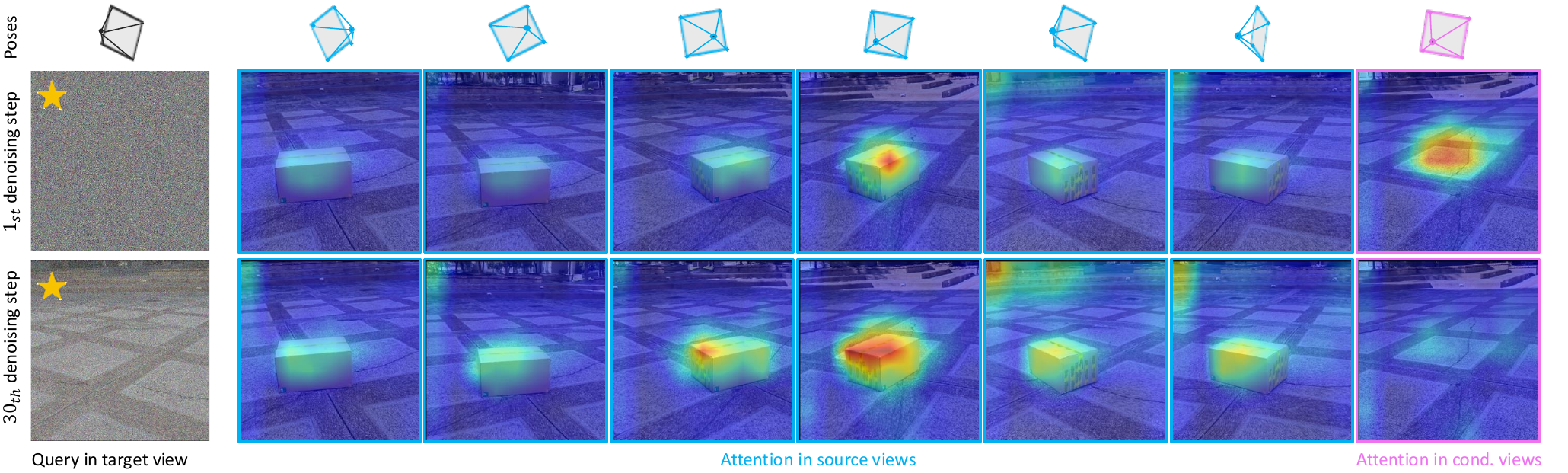}
    \caption{\textbf{Attention visualization of OmniNet Block during different denoising stages.} Omni-3DEdit can capture corresponding regions from source and conditional views based on the given camera poses. Pentagram is placed on the query regions.}
    \label{fig:attn_vis}
\end{figure*}

\begin{table*}[h!]
\centering
\resizebox{0.95\linewidth}{!}{
\begin{tabular}{@{}llll@{}}
\toprule
\textbf{Scene} & \textbf{Edit Instruction} & \textbf{Source Prompt} & \textbf{Target Prompt} \\ \midrule
Bear & ``Turn it into a black bear.'' & ``A bear in the park.'' & ``A black bear in the park.'' \\
Bear & ``Remove the bear.'' & ``A bear in the park.'' & ``A plain stone in the park.'' \\
Book & ``Remove the book + add an apple.'' & ``A table with a book.'' & ``A table with an apple.'' \\
Spinnerf\_1 & ``Remove the trash + add a poster.'' & ``A trash on the ground.'' & ``A poster.'' \\
Spinnerf\_2 & ``Remove the box under tree.'' & ``A box under the tree.'' & ``A tree.'' \\
CO3D\_Keyboard & ``Make the keyboard green + add a bottle.'' & ``A keyboard on the table.'' & ``A green keyboard and a bottle on the table.'' \\
Garden & ``Remove the vase.'' & ``A vase in the garden.'' & ``A wooden table in the garden.'' \\
Garden & ``Make the table blue.'' & ``A table.'' & ``A blue table.'' \\
Spinnerf\_7 & ``Remove the white object + add a plant.'' & ``A kettle on bench.'' & ``A plant on the bench.'' \\
Bicycle & ``Make the bicycle golden.'' & ``A bicycle on the glass.'' & ``A golden bicycle on the glass.'' \\
\bottomrule
\end{tabular}
}
\caption{\textbf{The scene and instruction prompts in the datasets.}}
\label{tab:prompts}
\end{table*}

\noindent \textbf{Training Settings.}
We train Omni-3DEdit using the AdamW~\cite{adamw} optimizer with $\beta_1=0.9$, $\beta_2=0.999$, and a weight decay of $0.01$. The training process utilizes a fixed learning rate of $1 \times 10^{-4}$ and a batch size of $32$ for a total of 4,000 training steps. Following the setting of SEVA~\cite{seva} training setup, we employ the EpsWeight loss strategy but omit weighting from camera distance. Furthermore, we adopt an SNR shift of $2.4$ to improve the signal-to-noise ratio scheduling. For classifier-free guidance (CFG), we apply a condition dropout rate (UCG rate) of $0.2$ during training and use a guidance scale of $1.2$ during inference.

\section{Attention Visualization}
\label{sec:add_exp}

To probe how different views are leveraged during denoising, we visualize the attention distribution across distinct denoising stages. As illustrated in Fig.~\ref{fig:attn_vis}, we randomly select a target view and analyze the attention map corresponding to its top-left region within the multi-view attention block.
At the initial denoising step ($1^{st}$ step), the query region predominantly attends to areas exhibiting significant discrepancies between the reference and source views, $a.k.a.$, regions of boxes present and absent. Notably, the source view aligned with the reference view's camera pose garners the highest attention. This facilitates model's interpretation of the intended task (3D removal) and the target object.
At the $30^{th}$  denoising step, the focused regions broaden beyond attending to the box's location across all source views, and it explicitly attends to regions in the original views that are geometrically coincident with the selected top-left position. This suggests that such source geometric textures are effectively preserved by Omni-3DEdit. These findings validate Omni-3DEdit's ability to utilize camera poses for resolving inter-view geometric relationships and precisely localizing relevant content from source views.

\section{Additional Test Data Details}
\label{sec:test_data_detail}

We present more details of our manually constructed cases in Tab.~\ref{tab:prompts}. These cases encompass standard appearance editing instructions to align with prior methodologies, such as `Turn it into a black bear'. Furthermore, we introduce more complex editing instructions, such as sequentially modifying an object's color before adding an additional object to the scene. As discussed in the main text, these composite instructions not only rely on the model's generalization capabilities but also necessitate high-fidelity results to support multi-round editing—a task that remains challenging for existing approaches. 

\begin{figure*}[h!]
    \centering
    \includegraphics[width=\linewidth]{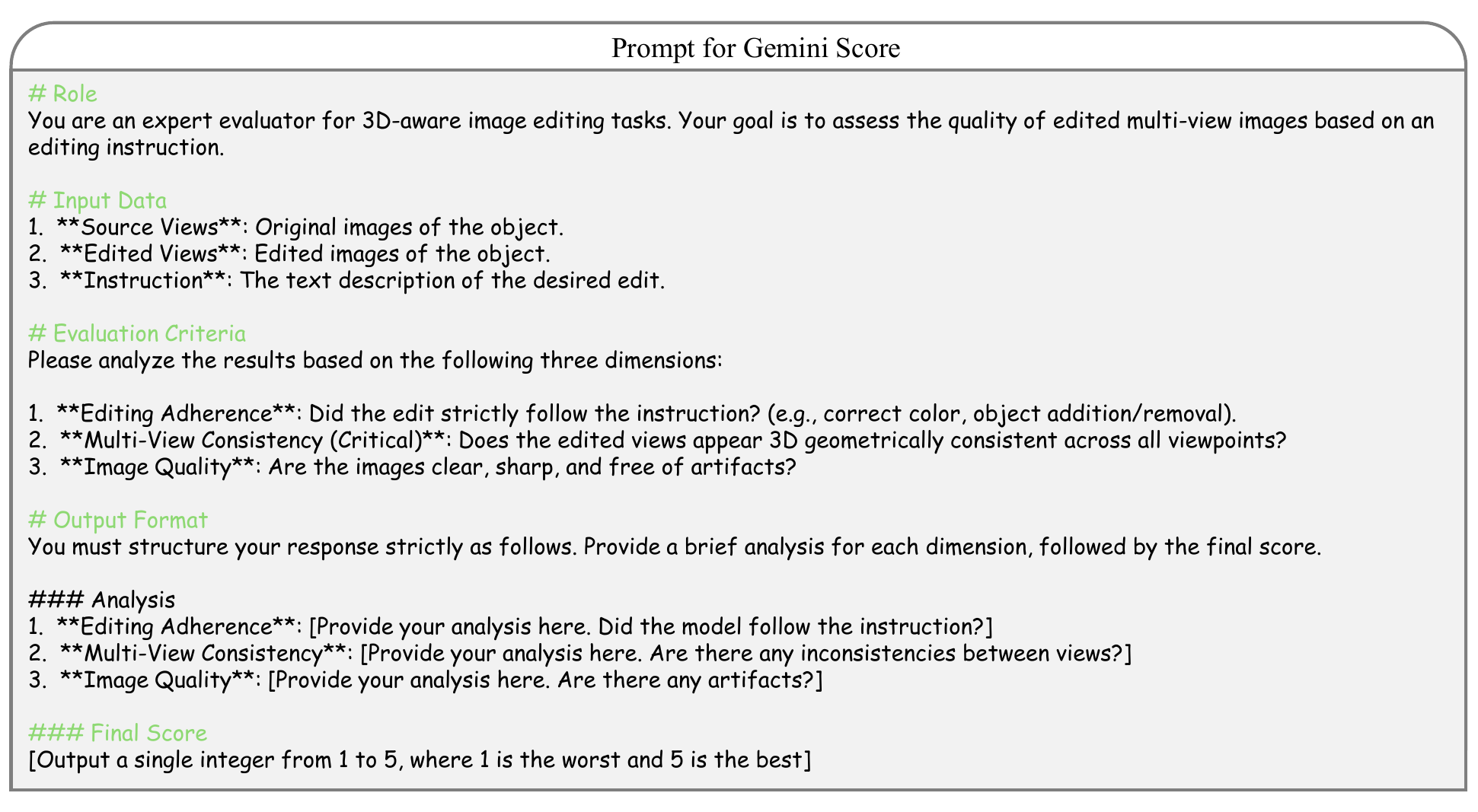}
    \caption{\textbf{System prompts for gemini score.}  }
    \label{fig:gemini_score}
\end{figure*}
\vspace{+1mm}

\noindent \textbf{Gemini Score.}
We employ evaluation criteria similar to those used in our quality filter, tasking Gemini-2.5 Pro~\cite{gemini} with assessing both the success of the multi-view editing and the maintenance of multi-view consistency. As shown in Fig.~\ref{fig:gemini_score}, we instruct Gemini to give a score from 1 to 5 by considering the editing instruction faithfulness, multi-view consistency, and visual quality comprehensively.

{
    \small
    \bibliographystyle{ieeenat_fullname}
    \bibliography{main}
}